\documentclass[final,dvipsnames]{colt2020} 

\usepackage{natbib}
\bibpunct{(}{)}{;}{a}{,}{,}

\usepackage{mathtools}
\usepackage{amsmath}
\usepackage{bbm}
\usepackage{amsfonts}
\usepackage{amssymb}
\let\vec\undefined
\usepackage{MnSymbol} 

\usepackage{xpatch}

\DeclarePairedDelimiter{\abs}{\lvert}{\rvert} %
\DeclarePairedDelimiter{\brk}{[}{]}
\DeclarePairedDelimiter{\crl}{\{}{\}}
\DeclarePairedDelimiter{\prn}{(}{)}

\DeclareMathOperator{\En}{\mathbb{E}}

\newcommand{\ls}{\ell}

\newcommand{\veps}{\varepsilon}

\newcommand{\ldef}{\vcentcolon=}

\newcommand{\wt}[1]{\widetilde{#1}}

\def\ddefloop#1{\ifx\ddefloop#1\else\ddef{#1}\expandafter\ddefloop\fi}
\def\ddef#1{\expandafter\def\csname bb#1\endcsname{\ensuremath{\mathbb{#1}}}}
\ddefloop ABCDEFGHIJKLMNOPQRSTUVWXYZ\ddefloop
\def\ddefloop#1{\ifx\ddefloop#1\else\ddef{#1}\expandafter\ddefloop\fi}
\def\ddef#1{\expandafter\def\csname b#1\endcsname{\ensuremath{\mathbf{#1}}}}
\ddefloop ABCDEFGHIJKLMNOPQRSTUVWXYZ\ddefloop
\def\ddef#1{\expandafter\def\csname c#1\endcsname{\ensuremath{\mathcal{#1}}}}
\ddefloop ABCDEFGHIJKLMNOPQRSTUVWXYZ\ddefloop
\def\ddef#1{\expandafter\def\csname h#1\endcsname{\ensuremath{\widehat{#1}}}}
\ddefloop ABCDEFGHIJKLMNOPQRSTUVWXYZ\ddefloop
\def\ddef#1{\expandafter\def\csname hc#1\endcsname{\ensuremath{\widehat{\mathcal{#1}}}}}
\ddefloop ABCDEFGHIJKLMNOPQRSTUVWXYZ\ddefloop
\def\ddef#1{\expandafter\def\csname t#1\endcsname{\ensuremath{\widetilde{#1}}}}
\ddefloop ABCDEFGHIJKLMNOPQRSTUVWXYZ\ddefloop
\def\ddef#1{\expandafter\def\csname tc#1\endcsname{\ensuremath{\widetilde{\mathcal{#1}}}}}
\ddefloop ABCDEFGHIJKLMNOPQRSTUVWXYZ\ddefloop

\usepackage[utf8]{inputenc} 
\usepackage[T1]{fontenc}    
\usepackage{url}            
\usepackage{booktabs}       
\usepackage{amsfonts}       
\usepackage{nicefrac}       
\usepackage{microtype}      
\usepackage{mkolar_definitions}
\usepackage{algorithm}
\usepackage[noend]{algorithmic}
\usepackage{wrapfig}

\usepackage{xspace}

\usepackage{enumitem}

\usepackage{breakcites}

\usepackage{hyphenat}
\usepackage{breakurl}

\def\Reg{\ensuremath{\text{\rm Reg}}}

\usepackage[scaled=.90]{helvet}

\newcommand{\expfour}{\textsf{Exp4}\xspace}
\newcommand{\adanormalhedge}{\textsf{AdaNormalHedge}\xspace}
\newcommand{\corral}{\textsf{Corral}\xspace}
\newcommand{\squint}{\textsf{Squint}\xspace}
\newcommand{\expthree}{\textsf{Exp3}\xspace}

\renewcommand{\Reg}{\mathrm{Reg}}

\usepackage{ifthen}

\newtheorem{inneropenproblem}{Open Problem}
\newenvironment{openproblem}[1]
  {\renewcommand\theinneropenproblem{#1}\inneropenproblem}
  {\endinneropenproblem}

  \newcommand{\mstar}{m_{\star}}
\newcommand{\fstar}{f_{\star}}

\newenvironment{packed_enum}{
  \begin{enumerate}
    \setlength{\itemsep}{1pt}
    \setlength{\parskip}{-1pt}
    \setlength{\parsep}{0pt}
}{\end{enumerate}}

\title[Open Problem: Model Selection for Contextual Bandits]{Open Problem: Model Selection for Contextual Bandits}
\usepackage{times}

\coltauthor{%
 \Name{Dylan J. Foster} \Email{dylanf@mit.edu}\\
 \addr Massachusetts Institute of Technology
 \AND
 \Name{Akshay Krishnamurthy} \Email{akshay@cs.umass.edu}\\
 \addr Microsoft Research NYC%
 \AND
  \Name{Haipeng Luo} \Email{haipengl@usc.edu}\\
 \addr University of Southern California%
}

\begin{document}

\maketitle

\begin{abstract}%
In statistical learning, algorithms for model selection allow the learner
to adapt to the complexity of the best hypothesis class in a
sequence. We ask whether similar guarantees are possible for
contextual bandit learning.
\end{abstract}


\section{Introduction}

Model selection is the fundamental statistical task of choosing a
hypothesis class using data, with statistical guarantees dating back
to Vapnik's structural risk minimization principle. Despite decades of
research on model selection for supervised learning and the ubiquity
of model selection procedures such as cross-validation in practice, very little is
known about model selection in interactive learning and reinforcement learning settings where
exploration is required. Focusing on contextual bandits, a simple reinforcement
learning setting, we ask: \emph{Can model selection guarantees be
  achieved in contextual bandit learning, where a learner must balance
  exploration and exploitation to make decisions online?}

\section{Problem Formulation}
We consider the adversarial contextual bandit setting
\citep{auer2002nonstochastic}. The setting is defined by a context
space $\Xcal$ and a finite action space $\Acal \ldef
\{1,\ldots,K\}$. The learner interacts with nature for $T$ rounds, where in
round $t$: (1) nature selects a context $x_t\in\cX$ and loss $\ls_t\in\brk*{0,1}^{\cA}$, (2) the learner
observes $x_t$ and chooses action $a_t$, and (3) the learner observes $\ls_t(a_t)$. We
allow for an adaptive adversary, so that $x_t$ and $\ls_t$ may depend on
$a_1,\ldots,a_{t-1}$. In the usual problem formulation, the learner is given a
policy class $\Pi\subset\prn*{\cX\to\cA}$, and the goal is to minimize
regret to $\Pi$:
\[
  \Reg(\Pi) \ldef \max_{\pi\in\Pi}\En\brk*{\sum_{t=1}^{T}\ls_t(a_t) - \sum_{t=1}^{T}\ls_t(\pi(x_t))}.
\]
When $\Pi$ is finite, the well-known \expfour algorithm
\citep{auer2002nonstochastic} achieves the optimal regret bound of
$O(\sqrt{KT\log\abs*{\Pi}})$.
\paragraph{The Model Selection Problem.} In the contextual bandit model selection problem, we assume that the policy class under consideration decomposes as a nested
  sequence:\footnote{It is also natural to consider infinite sequences
    of policy classes, but we restrict to finite sequences for simplicity.}
  \[
    \Pi_1\subset\Pi_2\subset\cdots\subset\Pi_M = \Pi.
  \]
The goal of the learner is to achieve low regret to all classes
  in the sequence simultaneously, with the regret to policy
  class $\Pi_m$ scaling only with $\log |\Pi_m|$. Intuitively,
  this provides a luckiness guarantee: if a good policy lies in a
  small policy class, the algorithm discovers this quickly.


    To motivate the precise guarantee we ask for, let us recall
    what is known in the simpler \emph{full-information} online
    learning setting, where the learner gets to see the entire loss
    vector $\ls_t$ at the end of each round. Here, the minimax rate is
    $O(\sqrt{T\log\abs*{\Pi}})$, and it can be shown
    (\cite{foster2015adaptive}; see also \cite{orabona2016coin}) that
    a variant of the exponential weights algorithm guarantees
      \begin{equation}
       \Reg(\Pi_m) 
        \leq{} O\prn*{
          \sqrt{T(\log\abs*{\Pi_{m}}+\log{}m)}
        },\quad\text{for all $m\in[M]$.}\label{eq:fullinfo}
      \end{equation}
      In other words, by paying a modest additive
      overhead of $\log m$, we can compete with all $M$ policy classes
      simultaneously.
      The most basic variant of our open problem asks whether the
      natural analogue of \pref{eq:fullinfo} can be attained for
      contextual bandits.
\begin{openproblem}{1a}
\label{op:adv}
Design a contextual bandit algorithm that for any sequence $\Pi_1\subset\Pi_2\subset\cdots\Pi_M$ ensures
\begin{equation}
          \max_{\pi\in\Pi_m}\En\brk*{\sum_{t=1}^{T}\ls_t(a_t) -
          \sum_{t=1}^{T}\ls_t(\pi(x_t))}
        \leq{} O\prn*{
          \sqrt{KT(\log\abs*{\Pi_{m}}+\log{}m)}
        },\quad\text{for all $m\in[M]$.}
  \label{eq:cb1}
\end{equation}
\end{openproblem}
We also welcome the following weaker guarantees.
\begin{openproblem}{1b}
\label{op:adv2}
Design a contextual bandit algorithm that for any sequence
$\Pi_1\subset\Pi_2\subset\cdots\Pi_M$ ensures either:
\begin{packed_enum}
\item $\Reg(\Pi_m) \leq O\prn*{\mathrm{poly}(K,M,\log\log\abs*{\Pi})\cdot{}\sqrt{T \log \abs*{\Pi_{m}}}}$ for all $m \in [M]$.
\item $\Reg(\Pi_m) \leq
  O\prn*{\mathrm{poly}(K,M,\log\log\abs*{\Pi})\cdot{}T^{\alpha}\log^{1-\alpha}\abs*{\Pi_m}}$
  for all $m \in [M]$, where $\alpha \in [\nicefrac{1}{2},1)$.
\end{packed_enum}
Alternatively, prove that no algorithm can achieve item 2 above for
any value $\alpha \in [\nicefrac{1}{2},1)$.
\end{openproblem}
The first item here differs from~\pref{op:adv} only in the dependence
on $K$, $M$, and $\log\log\abs*{\Pi}$
factors, which we do not believe
represent the most challenging aspect of the problem. The second item
is a further relaxation of the original guarantee. Here we simply ask
that the model selection algorithm has regret sublinear in $T$
whenever $\sqrt{T \log\abs*{\Pi_m}}$ is sublinear. In other words, if
policy class $\Pi_m$ is learnable on its own, the model selection
algorithm should have sublinear regret to it. To attain this behavior it is
essential that the exponents $\alpha$ and $1-\alpha$ sum to one.
Indeed, it is relatively easy to design algorithms with exponents that
do not sum to one,\footnote{For example we can attain regret $\sqrt{T}\cdot\log\abs*{\Pi_m}$ for all $m$ by running  \expfour with a
  particular prior over policies.} but we do not know of an algorithm satisfying
item 2 above for any $\alpha\in[1/2,1)$. This stands in contrast to
  other problems involving adaptivity and data-dependence in contextual
  bandits \citep{agarwal2017open}, where attaining adaptive guarantees
  with suboptimal dependence on $T$ is straightforward,
  and the primary challenge is to attain $\sqrt{T}$-type regret bounds. We also welcome a lower bound showing that this type of model
  selection guarantees is not possible for contextual bandits.





\paragraph{Stochastic Setting.}
The model selection problem for contextual bandits has yet to be
solved even for the stochastic setting, and even when the model is well-specified. Here, we assume: (1)
$\crl*{(x_t,\ls_t)}_{t=1}^{T}$ are drawn i.i.d. from a fixed
distribution $\cD$; (2) Each class $\Pi_m$ is induced by a class of
regression functions $\cF_m\subset{}(\cX\times\cA\to\brk*{0,1})$, in
the sense that $\Pi_m=\crl*{\pi_f\mid{}f\in\cF_m}$, where
$\pi_f(x)\ldef{}\argmin_{a\in\cA}f(x,a)$; (3) The problem is
realizable/well-specified in the sense that there exists index
$\mstar$ and regression function $\fstar\in\cF_{\mstar}$ such that
$\En\brk*{\ls(a)\mid{}x} = \fstar(x,a)$, for all $x,a$.  

 The final version of our open problem asks for a model selection
 guarantee when the problem is stochastic and well-specified. Note that
these assumptions imply that the optimal unconstrained policy (in
terms of expected loss) is $\pi_{\fstar}$. As such, here we only ask
for a regret bound against class $\Pi_{\mstar}$. From an
 algorithmic perspective, this is the easiest version of the problem.
\begin{openproblem}{2}
  \label{op:real}
   For some value $\alpha\in[1/2,1)$, design an algorithm for
   contextual bandits that for any sequence
   $\cF_1\subset\cF_2\subset\cdots\cF_{M} = \cF$, whenever data is stochastic and
   realizable, ensures
\begin{equation}
  \En\brk*{\sum_{t=1}^{T}\ls_t(a_t) -
          \sum_{t=1}^{T}\ls_t(\pi_{\fstar}(x_t))}
        \leq{} O\prn*{
          \mathrm{poly}(K,M,\log\log\abs*{\cF})\cdot{}T^{\alpha}\log^{1-\alpha}\abs*{\cF_{\mstar}}
        }.\label{eq:cb4}
      \end{equation}
      Alternatively, prove that no algorithm can achieve this guarantee for
any value of $\alpha \in [\nicefrac{1}{2},1)$.
\end{openproblem}

We offer \$300 for the first solution to either
\savehyperref{op:adv}{Open Problem 1} or \pref{op:real}.

\section{Challenges and Partial Progress}
\label{sec:challenges}
Many natural algorithmic strategies for model selection fail under
bandit feedback. These include (a) running \expfour over all policies
with a non-uniform prior adapted to the nested policy class structure,
(b) the \corral aggregation strategy~\citep{agarwal2017corralling}, and
(c) an adaptive version of the classical $\epsilon$-greedy
strategy~\citep{langford2008epoch}. These strategies all require
tuning parameters (e.g., the learning rate ) in terms of the class
index $m$ of interest, and naive tuning gives guarantees of the form
$\tilde{O}(T^{\alpha}\log^\beta\abs*{\Pi_{m}})$ for
$\alpha+\beta > 1$. Adaptive online learning algorithms like
\adanormalhedge~\citep{luo2015achieving} and \squint~\citep{koolen2015second} also fail because they do not adequately
handle bandit feedback.\footnote{Their regret bounds do not contain
  the usual ``local norm'' term used in the analysis of \expfour and
  other bandit algorithms.} We refer the reader
to~\citet{foster2019model} for more details on these strategies in the
context of model selection. The main point here is that model selection for contextual bandits appears to
require new algorithmic ideas, even when we are satisfied with weak
$O\prn{T^{\alpha}\log ^{1-\alpha}\abs{\Pi_{m}}}$-type
rates where $\alpha>1/2$.

In a recent paper \citep{foster2019model}, we showed that a guarantee of the
form~\pref{eq:cb4} \emph{is} \nohyphens{achievable} when $\cF_m$
consists of linear functions in $d_m$ dimensions, under distributional
assumptions on $\cD$. Our strategy was inspired by the fact that if the optimal loss $L^\star =
\EE\sbr{\ls(\pi_{\fstar}(x))}$ is known, one can test if a given class
$\cF_m$ contains the optimal policy by running a standard contextual
bandit algorithm and checking whether it substantially underperforms
relative to $L^{\star}$.
In our linear
setup, we showed that one can estimate a surrogate for the
optimal loss $L^\star$ at a ``sublinear'' rate, which allowed us to
run this testing strategy and achieve a guarantee akin
to~\pref{eq:cb4} with no prior information. However, we do not know
if this strategy can succeed beyond specialized settings where
sublinear loss estimation is possible. Along these lines, \citet{locatelli2018adaptivity} also observe that knowledge of $L^\star$ can enable adaptive
guarantees in Lipschitz bandits, where adaptivity is not possible in
the absence of such information (such lower bounds do not appear to carry over to the contextual case).

For (non-contextual) multi-armed bandits, several lower bounds
demonstrate that model selection is \emph{not} possible. 
\citet{lattimore2015pareto} shows that for multi-armed
bandits, if we want to ensure $O(\sqrt{T})$ regret against a single
fixed arm instead of the usual $O(\sqrt{KT})$ rate, we must incur
$\Omega(K\sqrt{T})$ regret to one of the remaining arms in the worst case. This precludes
a model selection guarantee of the form $\sqrt{T |\Acal_m|}$ for
nested action sets $\Acal_1 \subset \Acal_2 \subset \ldots$, which is
a natural analogue of \pref{eq:cb1} for bandits.\footnote{This does not preclude a guarantee of
the form \pref{eq:cb1}, however, since we pay for the
maximum number of actions.} Related lower bounds are also known
for Lipschitz
bandits~\citep{locatelli2018adaptivity,krishnamurthy2019contextual}. On
the positive side, \cite{chatterji2019osom} show that, with
distributional assumptions, it is possible to adapt between multi-armed
bandits and linear contextual bandits.





\section{Consequences and Connections to Other Problems}
\label{sec:consequences}
\paragraph{Switching Regret for Bandits.}
\newcommand{\astar}{a^{\star}}
In full-information online learning, algorithms for 
\emph{switching regret}
\citep{herbster1998tracking} simultaneously
ensure that for
all sequences of actions $\astar_1,\ldots,\astar_{T}$, $\En\brk*{\sum_{t=1}^{T}\ls_t(a_t) -
    \sum_{t=1}^{T}\ls_t(\astar_{t})}
  \leq{} O\prn*{
    \sqrt{S(\astar_{1:T})\cdot{}T}
    }$, where $S(\astar_{1:T})$ denotes the number of switches in the
    sequence. In the (non-contextual) multi-armed bandit setting, with no prior
    knowledge of the number of switches $S$, the best guarantee we are
    aware of is $      \En\brk*{\sum_{t=1}^{T}\ls_t(a_t) -
        \sum_{t=1}^{T}\ls_t(\astar_{t})}
      \leq{} O\prn*{
        \sqrt{S(\astar_{1:T})\cdot{}KT} + T^{3/4}
    }$
  which can be attained by combining the Bandits-over-Bandits strategy
  from \cite{cheung2019learning} with
  \expthree.\footnote{\cite{auer2002nonstochastic} achieves regret
    $\wt{O}(\sqrt{S\cdot{}KT})$, but only when a bound $S$ on
    the switches is known a-priori.} 
A solution to
  \pref{op:adv} would immediately yield a nearly-optimal switching
  regret bound of $\wt{O}(\sqrt{S(\astar_{1:T})\cdot{}KT})$ for
  bandits by choosing the $m$th policy class $\Pi_m$ to be the set  of
  all sequences $\astar_1,\ldots,\astar_{T}$ with at most $m$ switches.\footnote{Formally, this is accomplished by setting $\cX=\brk*{T}$ and
  $\pi(t)=\astar_t$.} Solving \pref{op:adv} would also lead to improvements in
switching regret for contextual bandits.

\paragraph{Second-Order Regret Bounds for Online Learning.}
Consider full-information online learning, and let $P_t$
denote the algorithm's distribution over policies at time $t$. An unresolved COLT 2016 open
problem of \cite{freund2016open} asks whether there exists an
algorithm for this setting with
regret at most $O\prn*{\sqrt{\sum_{t=1}^{T}\mathrm{Var}_{\pi\sim{}P_t}(\ls_t(\pi(x_t)))\cdot\log(1/\veps)
  }}$ against the top $\veps$-quantile of policies for all $\veps>0$
simultaneously. A slight strengthening of Freund's open problem asks
for the following bound:
\begin{equation}
  \sum_{t=1}^{T}\En_{\pi\sim{}P_t}\ls(\pi(x_t)) -
  \En_{\pi\sim{}Q}\ls_t(\pi(x_t))
  \leq{} O\prn*{\sqrt{\sum_{t=1}^{T}\mathrm{Var}_{\pi\sim{}P_t}(\ls_t(\pi(x_t)))\cdot\mathrm{KL}(Q\|P_1)
    }},\quad\forall{}Q\in\Delta_{\Pi}.
  \label{eq:pacbayes}
\end{equation}
\pref{eq:pacbayes} implies the weaker quantile bound by choosing $Q$ to be
uniform over the top $\veps$-fraction of policies and $P_1$ to be the
uniform distribution over all policies. While the $\log(1/\veps)$-type
quantile bound does not seem to imply \pref{eq:pacbayes} directly,
historically KL-based bounds have quickly followed quantile bounds
\citep{chaudhuri2009parameter,luo2015achieving,koolen2015second}.


The guarantee in \pref{eq:pacbayes} would immediately yield a positive
resolution to \pref{op:adv} via the following reduction: (1) Choose
$P_1(\pi)\propto\frac{1}{\abs*{\Pi_m}m^{2}}$ for all $\pi\in\Pi_m$; (2)
To handle bandit feedback, draw $a_t\sim{}p_t$ and feed importance weighted losses
$\hat{\ls}_t(a)\ldef{}\frac{\ls_t(a)}{p_t(a)}\ind\crl*{a_t=a}$ into the
full-information algorithm at each round, where
$p_t(a)\ldef{}\sum_{\pi\in \Pi}P_t(\pi)\ind\crl*{\pi(x_t)=a}$.  Conversely, a
lower bound showing that~\pref{eq:cb1} is not attainable would imply
that no full-information algorithm can achieve~\pref{eq:pacbayes},
which strongly suggests that the quantile bound in Freund's open
problem is also not attainable.


{\small \bibliography{refs}}
\appendix
%
%
%

\end{document}